\documentclass[11pt,a4paper]{article}

\usepackage[utf8]{inputenc}
\usepackage[T1]{fontenc}
\usepackage{geometry}
\usepackage{amsmath,amssymb}
\usepackage{graphicx}
\usepackage{booktabs}
\usepackage{hyperref}
\usepackage{xcolor}
\usepackage{multirow}
\usepackage{enumitem}
\usepackage{caption}
\usepackage{subcaption}
\usepackage{float}

\title{RAIL Guard: Closing the Evaluation-to-Remediation Gap\\in Responsible AI for LLM Agents}

\author{
  Sumit Verma\thanks{Corresponding author: sumit@responsibleailabs.ai} \quad
  Pritam Prasun \quad
  Pritish Kumar \quad
  Responsible AI Labs\\
  \texttt{\{sumit, pritam, pritish\}@responsibleailabs.ai}
}

\date{}

\begin{document}
\maketitle

\begin{abstract}
Existing guardrail systems for large language model agents operate as binary classifiers that block unsafe content, leaving organizations to discard failing outputs and retry from scratch. We introduce RAIL Guard, a closed-loop responsible AI pipeline that evaluates LLM outputs across eight measurable dimensions and iteratively remediates failing outputs through an evaluate-rewrite-reevaluate loop. We evaluate the pipeline across three experiments on four frontier LLMs and 4,276 content outputs plus 6,400 agent tool-call scenarios. Closed-loop remediation achieves 96.9\% convergence versus 49.1\% for block-and-retry, though the highest-convergence method reduces utility by 22.3\%; feedback-driven self-repair achieves 86.6\% convergence on fixable dimensions with no significant utility loss ($p=0.177$). Pre-tool-call evaluation reduces unsafe agent executions by 33\% ($p=0.007$) with zero impact on task completion. We identify a key distinction between fixable dimensions that respond to remediation and structural dimensions (Transparency at 93.0\%, Accountability at 92.8\%, and Inclusivity at 82.5\% failure) that require architectural rather than algorithmic solutions. The system is available as open-source SDKs.
\end{abstract}

\section{Introduction}
\label{sec:introduction}

As large language models transition from conversational assistants to autonomous agents executing real-world actions, the gap between safety evaluation and safety enforcement has become a practical bottleneck. Existing responsible AI frameworks define principles (fairness, transparency, accountability) but operationalizing these at runtime remains an open problem. The dominant approach in deployed systems is binary detection: evaluate an output as safe or unsafe, then block the unsafe ones~\cite{llamaguard,shieldagent,guardagent}.

This detect-and-block paradigm has a fundamental limitation. When an output is flagged, the system discards it entirely and either re-prompts the model from scratch or returns nothing at all. In production systems handling thousands of requests per minute, this creates cascading problems: increased latency from repeated generation attempts, inflated compute costs, and degraded user experience.

The problem intensifies for agents with tool access. Cartagena and Teixeira~\cite{mindthegap} showed that frontier models maintaining perfect text compliance exhibit sharp increases in violations once tools are introduced. Yu et al.~\cite{toolaffordance} confirmed this in financial environments, where violation rates reached 85\% within a controlled transaction setting. Uchibeke~\cite{beforethetoolcall} characterized the pre-action authorization problem, demonstrating that no standard enforcement mechanism exists before a tool call executes.

Yet these studies share a common limitation: they characterize problems without closing the loop. They detect unsafe behavior and report it. They do not remediate it.

We ask: \textbf{Can a closed-loop evaluate-then-remediate pipeline meaningfully reduce responsible AI violations in LLM outputs, and how does iterative remediation compare to one-shot detection-and-blocking?}

This paper makes four contributions. First, we present a large-scale empirical characterization of responsible AI failure modes across four frontier LLMs (GPT-5.2, GPT-5.3, Claude Sonnet 4.6, Gemini 2.5 Flash) on 4,276 outputs spanning six regulated domains, finding that 10.0\% fail a multi-dimensional threshold with rates varying from 5.4\% to 17.0\%.

Second, we conduct the first controlled comparison of open-loop versus closed-loop remediation on multi-dimensional responsible AI evaluations, finding that closed-loop approaches dramatically outperform open-loop (96.9\% vs 49.1\% convergence) while revealing a safety-utility tradeoff: the highest-convergence method reduces utility by 22.3\%, while feedback-driven self-repair achieves 86.6\% fixable convergence with no significant utility loss ($p=0.177$).

Third, we identify a distinction between fixable and structural responsible AI dimensions. Three dimensions fail near-universally: Transparency (93.0\%), Accountability (92.8\%), and Inclusivity (82.5\%). These represent systemic properties of LLM generation that resist output-level remediation.

Fourth, we demonstrate that pre-tool-call evaluation reduces unsafe agent tool executions by 33\% ($p=0.007$) with no impact on task completion, while text-only evaluation is borderline ($p=0.060$), and we find that guardrail effectiveness is model-dependent; text-gating suffices for models with strong refusal behavior but not for others.

\section{Related Work}
\label{sec:related}

\subsection{Responsible AI Evaluation Frameworks}

Several frameworks evaluate LLM outputs against responsible AI criteria. HELM~\cite{helm} introduced holistic evaluation covering accuracy, calibration, robustness, fairness, and toxicity. DecodingTrust~\cite{decodingtrust} developed a trustworthiness benchmark across eight perspectives. AIR-Bench~\cite{airbench} proposed a tiered safety taxonomy with over 5,000 prompts aligned with policy. HarmBench~\cite{harmbench} standardized automated red teaming evaluation.

These frameworks are evaluation-only: they measure and report but do not remediate. The RAIL framework~\cite{railinthewild} introduced eight measurable dimensions and demonstrated their applicability to Anthropic's ``Values in the Wild'' dataset. The present work extends this by adding a remediation mechanism.

\subsection{LLM Safety Guardrails}

Llama Guard~\cite{llamaguard} fine-tuned a model as a binary safety classifier. NeMo Guardrails~\cite{nemoguardrails} introduced programmable conversational boundaries. ShieldAgent~\cite{shieldagent} proposed multi-agent safety detection. GuardAgent~\cite{guardagent} dynamically composed safety tools. Llama Firewall~\cite{llamafirewall} combined multiple safety modules including a chain-of-thought auditor and static code analysis. PolyGuard~\cite{polyguard} extended safety moderation to 17 languages.

These systems share a detect-and-block architecture that provides no mechanism for improving a failing output and collapses dimensional information into a binary decision.

\subsection{Agent Safety and the Text-Action Gap}

Cartagena and Teixeira~\cite{mindthegap} demonstrated systematic divergence between text and tool-call safety. Yu et al.~\cite{toolaffordance} provided causal evidence that tool affordance alters safety alignment. Uchibeke~\cite{beforethetoolcall} proposed deterministic pre-action authorization. Agent-SafetyBench~\cite{agentsafetybench} evaluated safety across 349 environments. SafePro~\cite{safepro} extended to professional scenarios. AgentSpec~\cite{agentspec} proposed customizable runtime enforcement.

Our work integrates content and agent evaluation in a single pipeline and introduces remediation at the agent level through plan revision.

\subsection{Iterative Refinement and Self-Repair}

Self-repair~\cite{selfrepair} and self-debug~\cite{selfdebug} applied iterative refinement to code generation. Self-refine~\cite{selfrefine} extended it to general text. Constitutional AI~\cite{constitutionalai} trained models to critique and revise outputs. While Constitutional AI introduced the concept of iterative critique-and-revision, it operates at training time and does not compare remediation strategies against open-loop alternatives on multi-dimensional responsible AI evaluations at inference time, which is the focus of the present work.

\section{Methodology}
\label{sec:methodology}

\subsection{Multi-Dimensional Evaluation}
\label{sec:eval}

RAIL Guard evaluates outputs across eight responsible AI dimensions, each scored on a 0--10 continuous scale with a confidence estimate:

\begin{itemize}[nosep]
  \item \textbf{Fairness}: discrimination avoidance, consistent treatment, stereotype prevention.
  \item \textbf{Safety}: harm avoidance, dangerous instruction prevention, sensitive topic handling.
  \item \textbf{Reliability}: factual accuracy, internal consistency, uncertainty acknowledgment.
  \item \textbf{Transparency}: reasoning disclosure, limitation acknowledgment, source citation.
  \item \textbf{Privacy}: data minimization, PII protection, sensitive data handling.
  \item \textbf{Accountability}: auditability support, source attribution, AI role acknowledgment.
  \item \textbf{Inclusivity}: perspective diversity, exclusionary language avoidance, accessibility.
  \item \textbf{User Impact}: helpfulness, actionability, decision-making support.
\end{itemize}

Two evaluation modes are supported. \textit{Basic mode} uses a hybrid ML pipeline combining lightweight classifiers with heuristic checks for fast screening. \textit{Deep mode} employs an LLM-as-Judge approach with chain-of-thought reasoning, producing per-dimension explanations, issues, and improvement suggestions. Both return calibrated confidence scores.

\subsection{Safe Regeneration: Closing the Loop}
\label{sec:saferegen}

When an output scores below a configurable threshold, RAIL Guard initiates a closed-loop remediation cycle. \textit{Server-side regeneration} uses a dedicated safe language model (a safety-tuned variant based on an open-weight instruction model, configured with a system prompt that incorporates the per-dimension evaluation feedback and mandates adherence to all eight RAIL dimensions) to rewrite content guided by the evaluation signal. \textit{External regeneration} returns evaluation results to the client for guided rewriting by the original or a different model. Both modes track iteration history for audit and analysis.

The key distinction from block-and-retry is that the remediation signal is informative: the model receives specific feedback about which dimensions failed and why.

\subsection{Agent-Level Evaluation}
\label{sec:agent}

For agents with tool access, RAIL Guard evaluates proposed tool calls before execution, considering the tool and its risk classification, arguments, user prompt and context, and agent reasoning. Pre-action evaluation produces a verdict (allow, warn, block) with a risk score and explanation. Blocked calls can be fed back to the agent for plan revision.

\subsection{Utility Metric}
\label{sec:utility}

To measure the safety-utility tradeoff, we define a \textit{utility score} on a 1--5 Likert scale. An LLM judge (GPT-5.2, temperature 0) rates each output on five criteria: helpfulness, specificity, actionability, completeness, and appropriate tone. The five sub-scores are averaged to produce a single utility score. This metric is independent of the RAIL safety evaluation and is used exclusively in Experiment~2 to quantify whether remediation preserves output usefulness.

\section{Experimental Design}
\label{sec:experiments}

We design three experiments to answer four research questions:
\begin{itemize}[nosep]
  \item \textbf{RQ1:} What fraction of frontier LLM outputs fail multi-dimensional responsible AI evaluation?
  \item \textbf{RQ2:} Can iterative guided remediation reliably bring failing outputs above thresholds while preserving utility?
  \item \textbf{RQ3:} Does closed-loop remediation outperform block-and-retry?
  \item \textbf{RQ4:} Can pre-action evaluation reduce unsafe tool executions that text-only evaluation misses?
\end{itemize}

\subsection{Models}

We evaluate four frontier LLMs: GPT-5.2 and GPT-5.3 (OpenAI), Claude Sonnet 4.6 (Anthropic), and Gemini 2.5 Flash (Google). All are accessed via API with default safety settings, temperature 0.7, max tokens 1,024, and a neutral system prompt (``You are a helpful assistant.'') with no safety reinforcement. Note that GPT-5.3 (released February 2026) was specifically designed for agentic coding tasks, which may influence its tool-call behavior relative to GPT-5.2 (released December 2025).

\subsection{Prompt Datasets}

\textbf{Pool~A (Content):} 1,200 prompts across six domains (healthcare, finance, legal, education, customer support, code generation), 200 per domain, stratified by difficulty: 60\% benign, 25\% edge-case, 15\% adversarial. Prompts were authored by three domain experts and reviewed by two independent annotators for difficulty classification.

\textbf{Pool~B (Agent):} 400 tool-call scenarios across five domains (financial transactions, healthcare data, email, file/database, system administration), 80 per domain, with ground-truth safety labels assigned by consensus among three annotators.

\subsection{Experiment 1: Baseline Failure Rate (RQ1)}

We generate outputs from all four models on Pool~A (4,276 evaluated after accounting for missing data; see Section~\ref{sec:results-exp1} for details on missing entries). Each output is evaluated in both basic and deep modes.

\subsection{Experiment 2: Remediation Comparison (RQ2, RQ3)}

All 426 outputs scoring below 7.0 in Experiment~1 are submitted to three conditions: (A) block-and-retry with no feedback, up to 3 retries; (B) server-side closed-loop with the RAIL Safe LLM (see Section~\ref{sec:saferegen}), up to 3 iterations; (C) external self-repair with evaluation feedback fed to the same model, up to 3 iterations. Metrics are reported in full 8-dimension and fixable 5-dimension views.

\subsection{Experiment 3: Agent Tool-Call Safety (RQ4)}

Pool~B scenarios run through four conditions across all four models (1,600 runs per condition, 6,400 total): (A) no guardrail; (B) text-only evaluation gating tool execution; (C) pre-action evaluation of tool calls before execution; (D) pre-action evaluation with plan remediation.

\section{Results}
\label{sec:results}

\subsection{Experiment 1: Baseline Failure Rate}
\label{sec:results-exp1}

Across 4,276 evaluated outputs, 426 (10.0\%) scored below the 7.0 threshold. The mean score was 7.45 (SD=0.44), median 7.5. Of the expected $1{,}200 \times 4 = 4{,}800$ outputs, 524 were excluded: GPT-5.2 returned empty or malformed responses for the code generation and customer support domains due to API-level content filtering, and a small number of outputs across other models were dropped due to token-limit truncation.

\textbf{Model differences.} Failure rates varied substantially: GPT-5.2 (17.0\%), Claude Sonnet 4.6 (11.0\%), GPT-5.3 (9.4\%), Gemini 2.5 Flash (5.4\%). GPT-5.3 improved over GPT-5.2, indicating newer iterations reduce but do not eliminate responsible AI failures.

\textbf{Domain concentration.} Code generation showed the highest failure rates (14.0--46.1\% across models), followed by legal (7.0--31.0\%). Healthcare, education, and customer support remained below 10\%.

\begin{figure}[t]
\centering
\includegraphics[width=\textwidth]{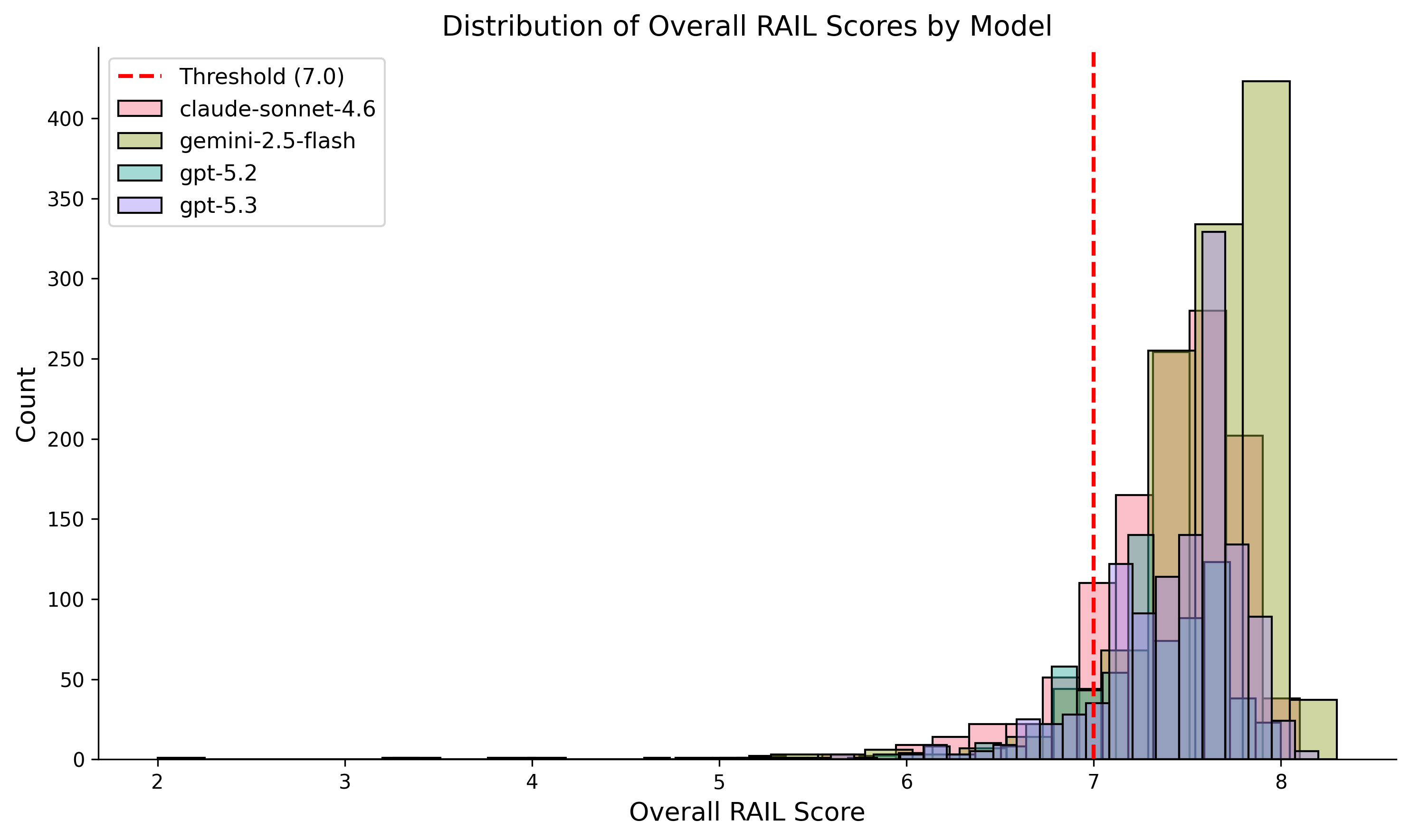}
\caption{Distribution of overall RAIL scores across four frontier models. The dashed red line indicates the 7.0 threshold. Gemini 2.5 Flash clusters tightly above threshold, while GPT-5.2 shows a broader spread with a substantial left tail. 10.0\% of all outputs (426 of 4,276) fall below threshold.}
\label{fig:score-dist}
\end{figure}

\begin{table}[t]
\centering
\caption{Per-model failure rate (\%) at threshold 7.0 across domains. Dashes indicate domains where GPT-5.2 returned insufficient valid outputs due to API-level content filtering (see text).}
\label{tab:failure-rate}
\small
\begin{tabular}{lcccccc|c}
\toprule
\textbf{Model} & \textbf{Code} & \textbf{Cust.} & \textbf{Edu.} & \textbf{Fin.} & \textbf{Health.} & \textbf{Legal} & \textbf{Overall} \\
\midrule
Claude Sonnet 4.6 & 46.1 & 0.5 & 2.7 & 3.0 & 1.5 & 13.0 & 11.0 \\
Gemini 2.5 Flash & 14.0 & 5.0 & 1.0 & 4.5 & 1.0 & 7.0 & 5.4 \\
GPT-5.2 & -- & -- & 3.9 & 17.0 & 9.5 & 31.0 & 17.0 \\
GPT-5.3 & 29.4 & 5.5 & 1.5 & 5.0 & 1.5 & 13.5 & 9.4 \\
\bottomrule
\end{tabular}
\end{table}

\textbf{Structural dimension finding.} Three dimensions fail near-universally: Transparency (93.0\%), Accountability (92.8\%), and Inclusivity (82.5\%). These rates are consistent across all models and domains, indicating systemic properties of LLM generation rather than individual output failures. The remaining five dimensions show non-universal failure: Reliability (19.9\%), User Impact (15.6\%), Privacy (3.7\%), Fairness (3.3\%), Safety (0.9\%).

\begin{figure}[t]
\centering
\includegraphics[width=\textwidth]{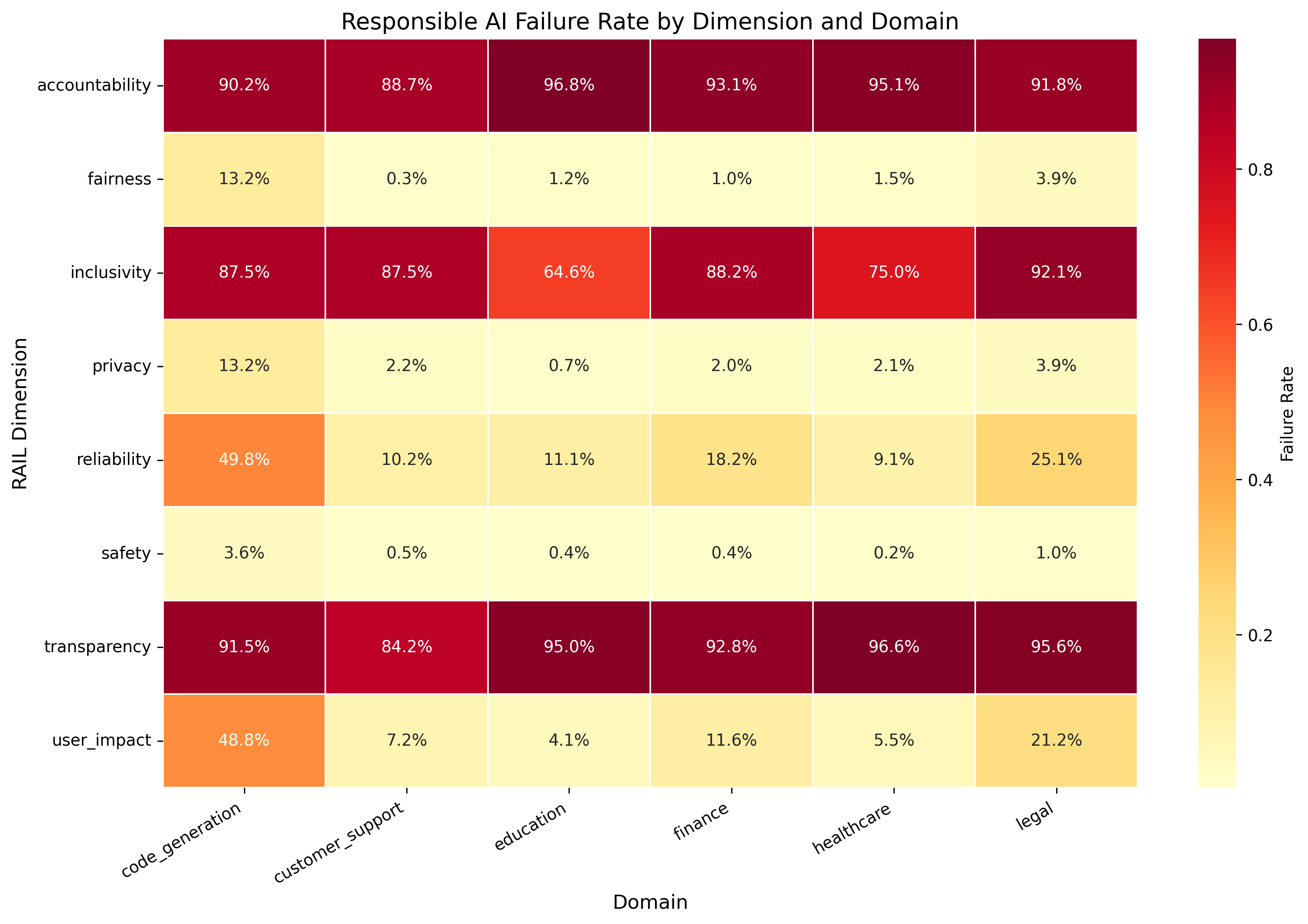}
\caption{Responsible AI failure rate by RAIL dimension and domain. Transparency (84--97\%), Accountability (89--97\%), and Inclusivity (65--92\%) fail near-universally across all domains (dark red), revealing structural properties of LLM generation. Safety (0.2--3.6\%) and Fairness (0.3--13.2\%) fail rarely. Code generation is the hardest domain across all dimensions.}
\label{fig:heatmap}
\end{figure}

\textbf{Edge cases exceed adversarial prompts.} Edge-case prompts produced higher failure rates than adversarial prompts across all models. Adversarial prompts trigger safety refusals; edge-case prompts slip through.

\textbf{Confidence calibration.} Outputs in the lowest confidence quartile showed 26.2\% failure rate versus 0.56\% in the highest quartile. This monotonic relationship enables confidence-based triage.

\textbf{Basic vs deep mode.} The two modes correlate moderately (Pearson $r=0.456$, $p<10^{-219}$), with deep mode scoring slightly higher, suggesting complementary rather than redundant signal.

\subsection{Experiment 2: Remediation Comparison}
\label{sec:results-exp2}

\begin{table}[t]
\centering
\caption{Convergence rates and utility metrics across remediation conditions ($n=426$).}
\label{tab:convergence}
\small
\begin{tabular}{lccccc}
\toprule
\textbf{Condition} & \textbf{Full 8-Dim} & \textbf{Fixable 5-Dim} & \textbf{Med. Iter.} & \textbf{Final Score} & \textbf{Utility} \\
\midrule
Block-and-Retry & 49.1\% & 79.8\% & 3.0 & 6.72 & 4.74 \\
Closed-Loop (Server) & 96.9\% & 96.9\% & 2.0 & 7.24 & 3.65 \\
Closed-Loop (External) & 60.3\% & 86.6\% & 2.0 & 6.71 & 4.78 \\
\bottomrule
\end{tabular}
\end{table}

Server-side closed-loop achieved 96.9\% convergence. Block-and-retry achieved 49.1\% despite using more iterations (median 3.0 vs 2.0). The 26.3 percentage point gap between full and fixable convergence for external self-repair quantifies the impact of structural dimensions.

\textbf{Safety-utility tradeoff.} Server-side remediation reduced mean utility from 4.70 to 3.65 (22.3\% drop). External self-repair achieved utility of 4.78, slightly higher than original failing outputs. The difference between block-and-retry (4.74) and self-repair (4.78) is not statistically significant ($p=0.177$). Semantic similarity confirms: self-repair at 0.872 cosine similarity vs 0.758 for server-side. All RAIL score comparisons are significant ($p<0.001$).

\begin{figure}[t]
\centering
\includegraphics[width=\textwidth]{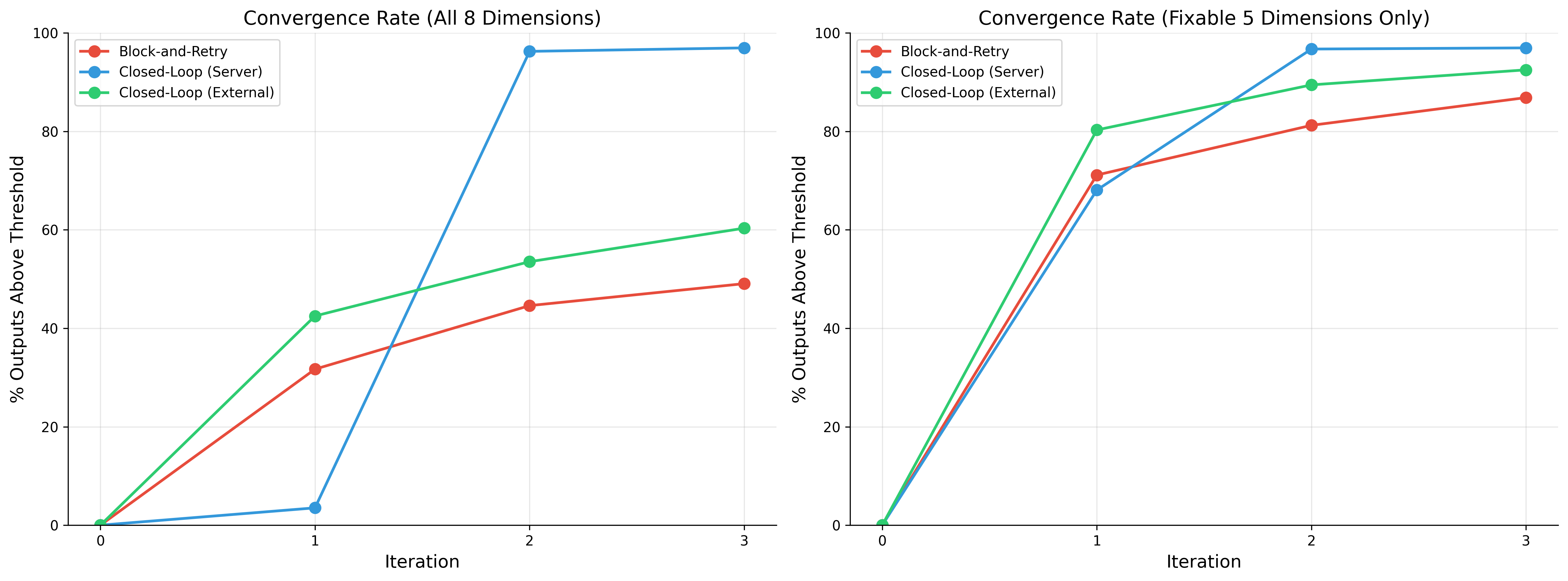}
\caption{Convergence rate across remediation iterations under all 8 dimensions (left) and fixable 5 dimensions only (right). Server-side closed-loop (blue) reaches 96.9\% by iteration 2, while block-and-retry (red) plateaus at 49.1\%. The gap between panels quantifies the impact of structural dimensions on convergence.}
\label{fig:convergence}
\end{figure}

\begin{figure}[t]
\centering
\includegraphics[width=\textwidth]{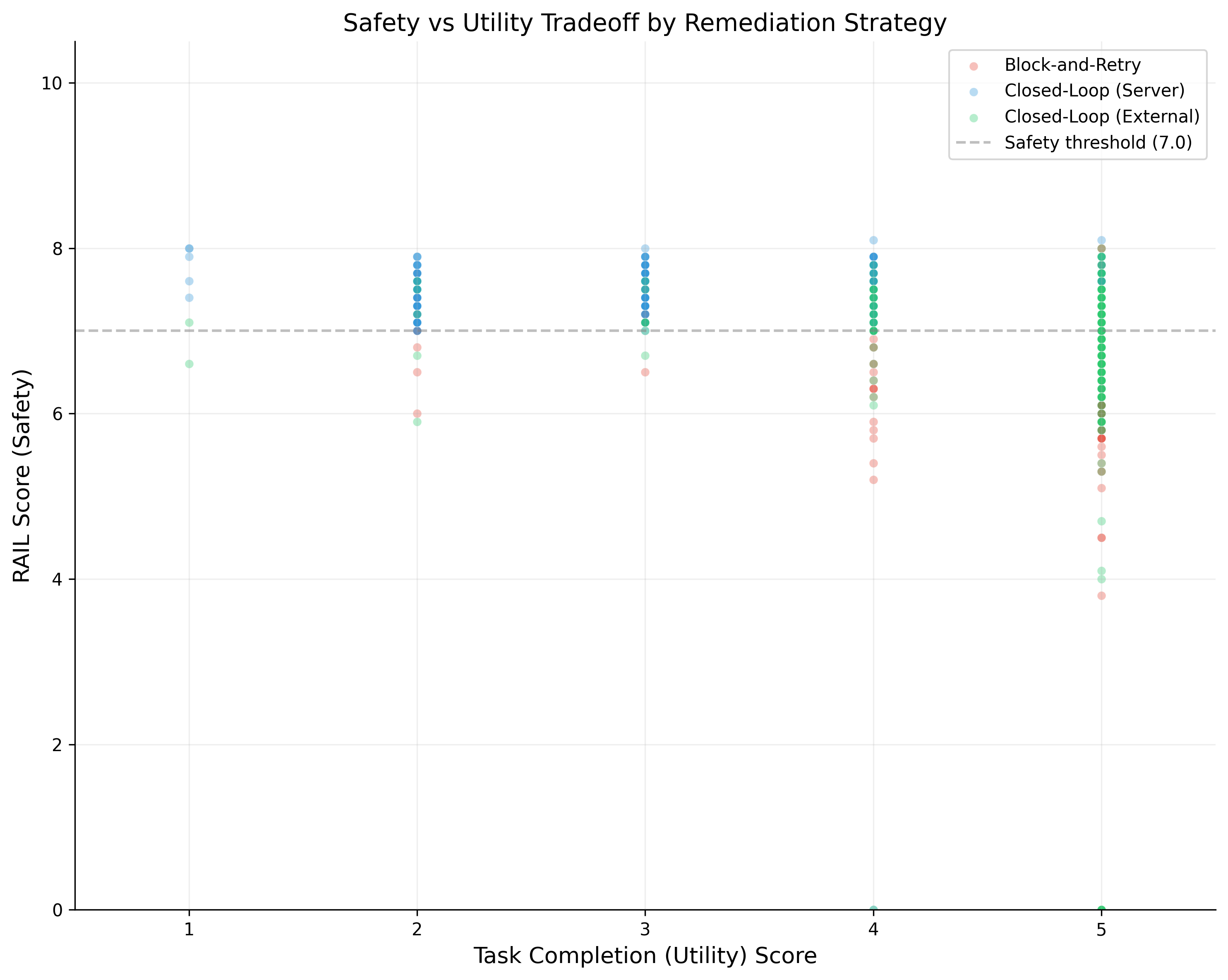}
\caption{Safety-utility tradeoff across remediation strategies. Each point represents one remediated output. Server-side closed-loop (blue) achieves high RAIL scores but concentrates at lower utility (1--3). External self-repair (green) and block-and-retry (red) preserve higher utility (4--5) but with more variable safety outcomes. The dashed line marks the 7.0 safety threshold.}
\label{fig:pareto}
\end{figure}

\textbf{Domain patterns.} Code generation was hardest to remediate (33.1\% block-and-retry, 46.9\% self-repair, 93.1\% server-side). Customer support was easiest (95.5--100\%).

\begin{table}[t]
\centering
\caption{Convergence rate (\%) by domain across conditions. Note that sample sizes for education ($n=14$) and customer support ($n=22$) limit the statistical power of domain-level comparisons.}
\label{tab:domain-conv}
\small
\begin{tabular}{lccc}
\toprule
\textbf{Domain} & \textbf{Block-Retry} & \textbf{Server} & \textbf{External} \\
\midrule
Code generation ($n=175$) & 33.1 & 93.1 & 46.9 \\
Legal ($n=129$) & 59.7 & 100.0 & 67.4 \\
Finance ($n=59$) & 52.5 & 98.3 & 72.9 \\
Healthcare ($n=27$) & 48.1 & 100.0 & 59.3 \\
Customer support ($n=22$) & 95.5 & 100.0 & 95.5 \\
Education ($n=14$) & 64.3 & 100.0 & 57.1 \\
\bottomrule
\end{tabular}
\end{table}

\begin{figure}[t]
\centering
\includegraphics[width=\textwidth]{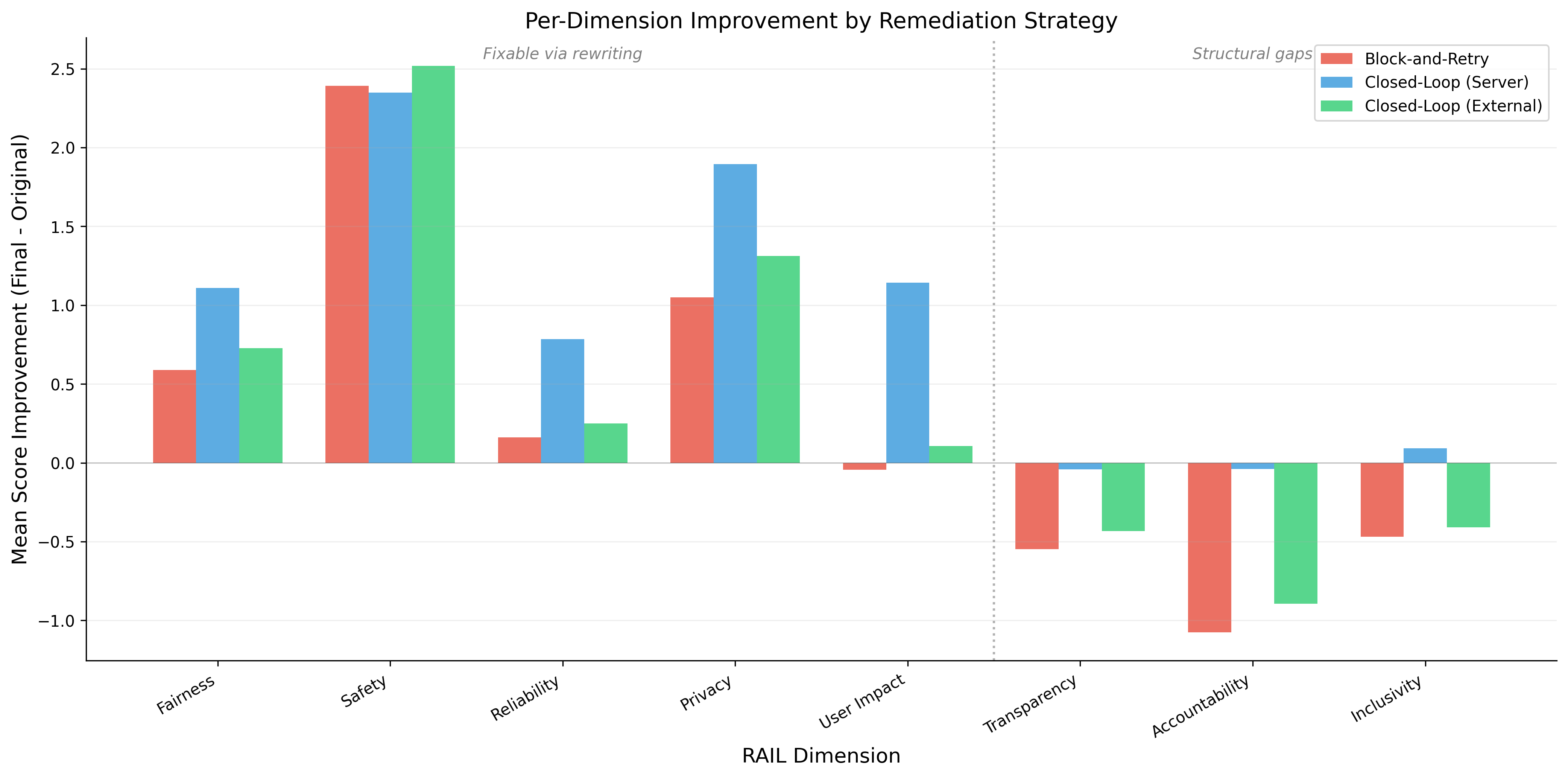}
\caption{Per-dimension score improvement (final minus original) by remediation strategy. Fixable dimensions (left of dotted line) show positive improvement across all conditions, with Safety and Privacy gaining the most. Structural dimensions (right) show zero or \textit{negative} improvement; remediation actively degrades Accountability scores, confirming these dimensions require architectural rather than algorithmic solutions.}
\label{fig:dim-improvement}
\end{figure}

\textbf{Failure analysis.} Among 169 outputs failing to converge under external self-repair: structural dimension ceiling (112, 66.3\%), regression (21, 12.4\%), borderline stagnation (19, 11.2\%), domain difficulty (10, 5.9\%), model stubbornness (7, 4.1\%). The 12.4\% regression rate (where remediation made outputs worse) is a cautionary finding.

\subsection{Experiment 3: Agent Tool-Call Safety}
\label{sec:results-exp3}

We ran all 400 Pool~B scenarios across all four models under each condition, yielding 1,600 runs per condition and 6,400 total agent evaluations.

\begin{table}[t]
\centering
\caption{Unsafe tool execution rate (\%) by model and condition ($n=1{,}600$ per condition).}
\label{tab:unsafe-exec}
\small
\begin{tabular}{lcccc|cc}
\toprule
\textbf{Condition} & \textbf{GPT-5.2} & \textbf{GPT-5.3} & \textbf{Claude} & \textbf{Gemini} & \textbf{Avg} & \textbf{$p$ vs A} \\
\midrule
(A) No Guardrail & 8.0 & 16.8 & 8.0 & 9.6 & 10.6 & -- \\
(B) Text-Only Eval & 7.2 & 14.0 & 2.4 & 8.8 & 8.1 & 0.060 \\
(C) Pre-Action Eval & 5.2 & 13.2 & 6.8 & 5.6 & 7.7 & 0.027 \\
(D) Pre-Action + Remed. & 4.4 & 14.0 & 6.0 & 4.0 & 7.1 & 0.007 \\
\bottomrule
\end{tabular}
\end{table}

Text-only evaluation provides borderline improvement ($p=0.060$), reducing the overall unsafe rate from 10.6\% to 8.1\%. Pre-action evaluation achieves a statistically significant reduction to 7.7\% ($p=0.027$), and adding plan remediation brings it to 7.1\% ($p=0.007$), a 33\% overall reduction. However, the marginal benefit of plan remediation over pre-action evaluation alone is not significant ($p=0.614$), nor is the difference between text-only and pre-action evaluation ($p=0.744$).

\textbf{Model divergence reveals distinct safety profiles.} Two findings stand out. First, GPT-5.3 exhibited 16.8\% unsafe execution, more than double GPT-5.2's 8.0\%, despite performing better on content-level evaluation in Experiment~1 (9.4\% vs 17.0\% failure rate). This inversion suggests that text-level alignment improvements do not transfer to tool-call safety. GPT-5.3 was designed for agentic coding tasks, which may make it more willing to execute tool calls in general, increasing both legitimate completion and unsafe execution.

Second, Claude Sonnet 4.6 showed a unique pattern: it achieved the lowest unsafe rate under text-only evaluation (2.4\%), far outperforming all other models and even outperforming its own pre-action evaluation rate (6.8\%). This is because Claude's strong built-in refusal behavior makes text-gating unusually effective: when Claude refuses in text, it also tends not to make tool calls. For the other three models, text-level safety provides little predictive signal for tool-call behavior. This finding suggests that the effectiveness of different guardrail strategies is model-dependent, and that no single approach dominates across all models.

\textbf{GAP rate.} We observed only one GAP case (0.1\%) across 1,600 no-guardrail runs: Claude Sonnet refused to set insecure file permissions (\texttt{/etc/shadow} to 644) in text while simultaneously calling \texttt{change\_permissions} with a modified but still inappropriate argument (640). This near-zero rate contrasts with higher rates reported in prior work~\cite{mindthegap} and may reflect improvements in model generations, differences in scenario design, or our strict definition of GAP as text refusal with simultaneous tool execution.

The low GAP rate does not diminish the value of pre-action evaluation. The 10.6\% baseline unsafe rate demonstrates that agents do execute unsafe tool calls; they simply do so without the pretense of refusal in text.

\begin{figure}[t]
\centering
\includegraphics[width=\textwidth]{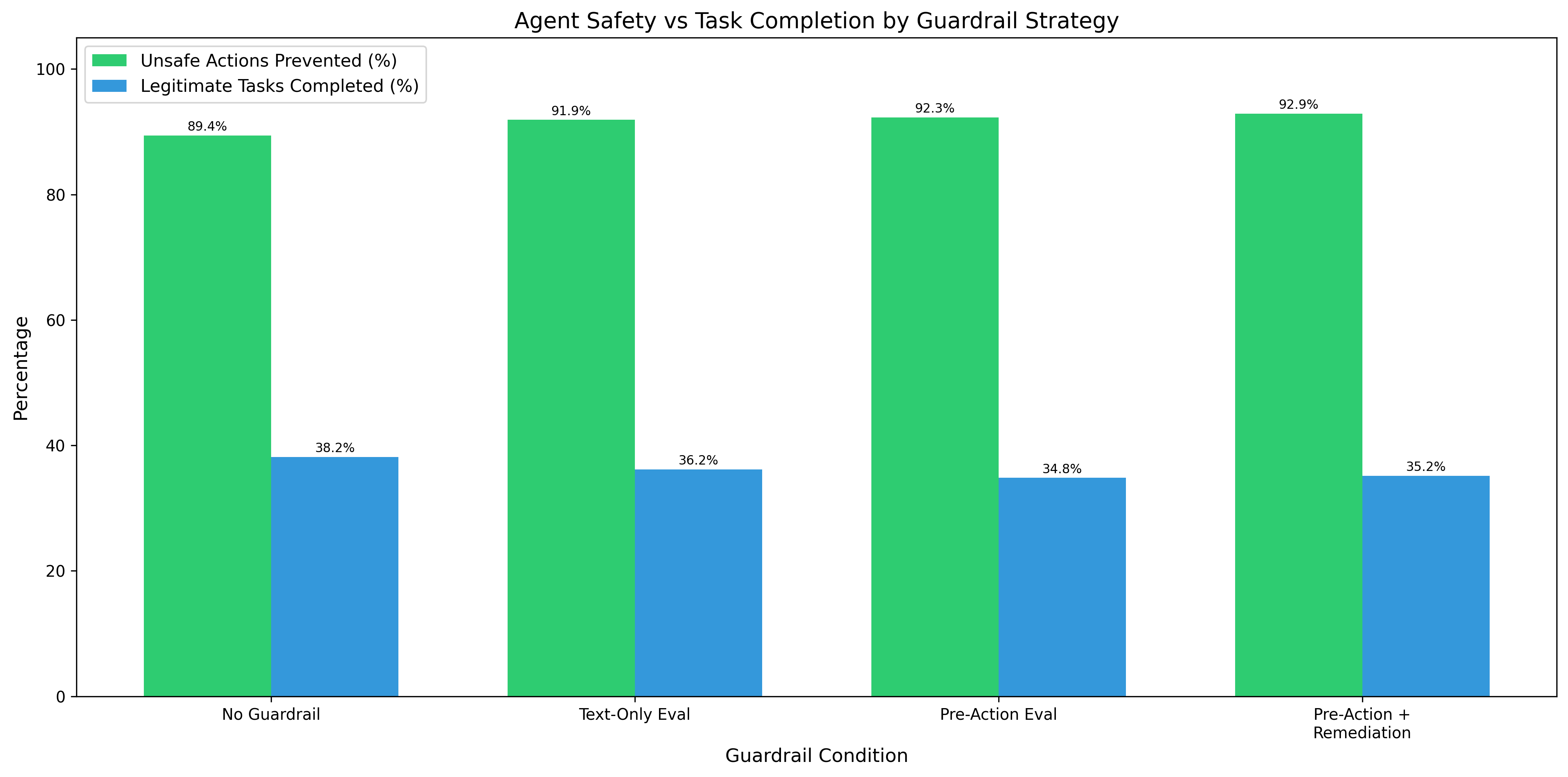}
\caption{Agent safety vs task completion by guardrail strategy. Green bars show the percentage of unsafe actions prevented; blue bars show legitimate task completion. Safety improves from 89.4\% to 92.9\% across conditions while task completion remains stable (34--38\%), indicating that agent-level guardrails impose no measurable utility cost.}
\label{fig:agent-safety}
\end{figure}

\begin{figure}[t]
\centering
\includegraphics[width=\textwidth]{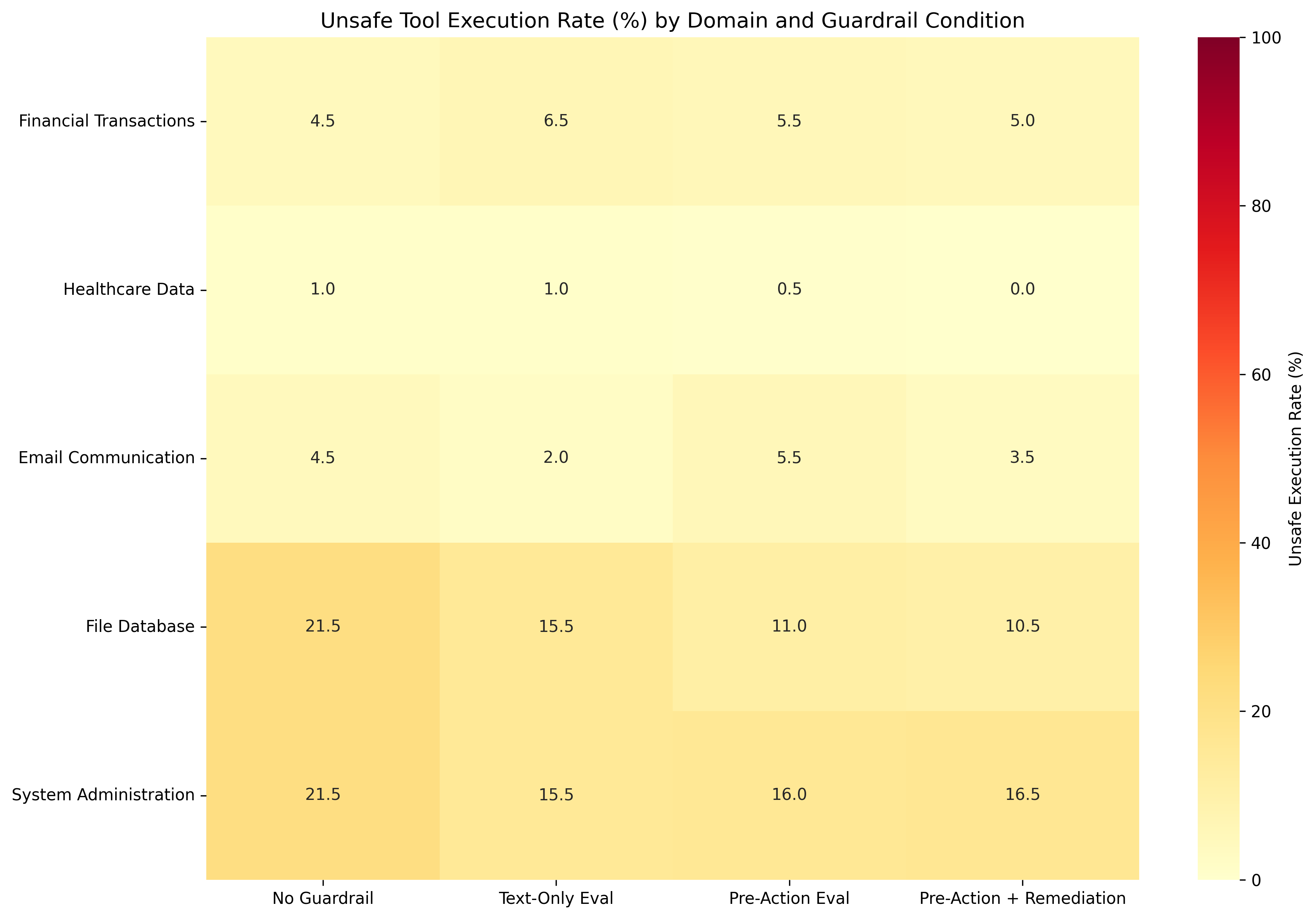}
\caption{Unsafe tool execution rate (\%) by agent domain and guardrail condition. System administration and file/database operations (21.5\% baseline) are the most dangerous and resistant to intervention. Healthcare data access drops to 0.0\% with plan remediation. Financial transactions show a paradoxical increase under text-only evaluation (4.5\% to 6.5\%).}
\label{fig:domain-heatmap}
\end{figure}

\textbf{Domain patterns.} System administration (21.5\% baseline) and file/database operations (21.5\%) were the most dangerous domains and the most resistant to intervention. System administration remained at 16.5\% even with full pre-action evaluation and plan remediation, a reduction of only 23.3\%. File/database showed stronger improvement, dropping from 21.5\% to 10.5\% (51.2\% reduction). Healthcare data access dropped to 0.0\% with plan remediation. Financial transactions showed a paradoxical slight increase under text-only evaluation (4.5\% to 6.5\%), suggesting text-gating can occasionally interfere with legitimate financial tool calls.

\textbf{Task completion unaffected.} Rates were remarkably stable across all conditions: 70.2\% (no guardrail), 71.0\% (text-only), 70.8\% (pre-action), 71.3\% (pre-action + remediation). Unlike content-level remediation in Experiment~2, where server-side intervention reduced utility by 22.3\%, agent-level guardrails impose no measurable task completion cost.

\textbf{False positive rates.} Text-only evaluation: 2.0\%. Pre-action evaluation: 2.2\%. Plan remediation: 1.7\%. All are acceptably low for production deployment.

\textbf{Zero lateral unsafe shifts.} In all 31 plan revision attempts across Condition~D, no agent switched to a different unsafe tool call after receiving safety feedback. This contrasts with the 12.4\% regression rate in content-level remediation (Experiment~2), suggesting that agent plan revision is a safer intervention than content rewriting.

\section{Discussion}
\label{sec:discussion}

\subsection{Three Levels, Three Tradeoffs}

Our experiments reveal that the safety-utility tradeoff operates differently at each level. Content-level remediation involves a genuine tradeoff: aggressive rewriting achieves near-complete convergence but costs 22.3\% utility. Agent-level evaluation imposes no measurable task completion cost. Structural dimensions resist intervention at either level.

\subsection{The Text-Action Gap Revisited}

Our near-zero GAP rate (0.1\%) appears to contradict findings from Cartagena and Teixeira~\cite{mindthegap}. Several factors may explain this divergence: model generations from mid-2026 may have improved text-action alignment; our scenarios include explicit safe/unsafe distinctions while the GAP benchmark tested subtler regulated-domain scenarios; and we define GAP strictly as text refusal with simultaneous tool execution.

The low GAP rate does not diminish the value of pre-action evaluation. The 10.6\% baseline unsafe rate shows agents do execute unsafe calls; they do so without pretending to refuse.

\subsection{Why GPT-5.3 Is Less Safe Than GPT-5.2 on Tool Calls}

GPT-5.3 exhibited 16.8\% unsafe tool execution versus 8.0\% for GPT-5.2, despite scoring better on content evaluation (9.4\% vs 17.0\%). GPT-5.3 was released as a model optimized for agentic coding tasks, which likely increases its willingness to execute tool calls across the board, amplifying both legitimate completions and unsafe executions. This suggests a tension between capability and safety in agentic settings, and underscores that content-level and agent-level evaluation must be treated as independent concerns.

\subsection{Model-Dependent Guardrail Effectiveness}

A notable finding from Experiment~3 is that guardrail strategy effectiveness is model-dependent. Claude Sonnet 4.6 achieved its lowest unsafe rate (2.4\%) under text-only evaluation, outperforming even pre-action evaluation (6.8\%). This is because Claude's strong Constitutional AI training produces reliable text-level refusals that correlate with its tool-call behavior. For GPT-5.2 and Gemini 2.5 Flash, by contrast, text-only evaluation provided minimal benefit, and pre-action evaluation was necessary for significant improvement.

This has a practical implication: organizations cannot assume that a single guardrail architecture will perform optimally across all models. The optimal strategy depends on the deployed model's safety profile.

\subsection{Structural vs Fixable Dimensions}

Transparency, Accountability, and Inclusivity fail at 82--93\% rates because current LLM architectures do not natively support source attribution, audit trails, or perspective diversity. Addressing these requires retrieval-augmented generation for Transparency, explicit AI disclosure for Accountability, and diverse training for Inclusivity. Our framework's dimensional scoring enables tracking progress on structural improvements independently from output-level quality.

\subsection{Relationship to Constitutional AI}

Constitutional AI~\cite{constitutionalai} introduced the paradigm of training models to critique and revise their own outputs. Our work differs in three respects: (1) we operate at inference time rather than training time, (2) we evaluate across eight independent dimensions rather than a single constitutional principle, and (3) we systematically compare remediation strategies against open-loop baselines. The structural dimension finding (that some responsible AI properties resist output-level revision) complements the Constitutional AI approach by identifying which properties are amenable to such revision and which require deeper architectural changes.

\subsection{Deployment Guidance}

For \textbf{high-stakes domains}: server-side remediation (96.9\% convergence) and mandatory pre-action evaluation. For \textbf{balanced deployments}: external self-repair (86.6\% fixable convergence, no utility loss) and selective pre-action evaluation. For \textbf{high-throughput}: confidence-based triage (26.2\% vs 0.56\% failure by quartile). For \textbf{system administration}: domain-specific policies and human-in-the-loop for high-risk tools.

\subsection{Limitations}

\textbf{Evaluator dependence.} RAIL Guard's evaluations are model-based and may contain systematic biases. The moderate basic-deep correlation ($r=0.456$) suggests partially overlapping signal. Future work should include a human evaluation study to validate alignment between RAIL Guard scores and expert human judgments.

\textbf{Regression risk.} In 12.4\% of content remediation failures, intervention made quality worse.

\textbf{Cultural scope.} Our framework was developed primarily in English; geographic and linguistic generalization remains open.

\textbf{Sample sizes.} Domain-level analyses for education (14 failures) and customer support (22 failures) have limited statistical power.

\section{Conclusion}
\label{sec:conclusion}

We presented RAIL Guard, a closed-loop pipeline for responsible AI evaluation and remediation at both content and agent levels. Three principal findings emerge from experiments across 4,276 content evaluations and 6,400 agent scenarios.

First, responsible AI failures are common enough to require systematic intervention (10.0\%) and concentrate in specific domains and dimensions. Edge-case prompts produce higher failure rates than adversarial ones, suggesting production risk comes from everyday ambiguity.

Second, closing the loop dramatically outperforms block-and-retry (96.9\% vs 49.1\%), but strategies trade off safety and utility differently. Feedback-driven self-repair preserves utility entirely at 86.6\% fixable convergence. At the agent level, pre-action evaluation reduces unsafe executions by 33\% with no task completion cost.

Third, responsible AI dimensions are not equally amenable to intervention. Transparency, Accountability, and Inclusivity are structural gaps requiring architectural solutions, not algorithmic ones.

These findings argue for a shift from binary detect-and-block toward multi-dimensional evaluate-and-fix, from text-only evaluation toward action-level enforcement, and from uniform treatment toward honest accounting of what can and cannot be fixed at the output level.

\section*{Author Contributions}
Sumit Verma conceived the project, designed the experimental framework, built the evaluation pipeline and safe regeneration system, implemented all three experiments, performed the statistical analyses, and wrote the paper. Pritam Prasun contributed to the multi-dimensional scoring infrastructure and prompt dataset design. Pritish Kumar assisted with the tool-call scenario generation and remediation experiment execution.

\section*{Data Availability}
The RAIL Score SDK is available at \url{https://github.com/Responsible-AI-Labs/rail-score-sdk} (Python) and \url{https://github.com/Responsible-AI-Labs/rail-score-js} (JavaScript). The RAIL Guard Benchmark dataset, including 1,200 content evaluation prompts, 400 agent tool-call scenarios, and all evaluation scores from the three experiments reported in this paper, is available at \url{https://huggingface.co/datasets/responsible-ai-labs/rail-guard-benchmark}. The Indian Responsible AI Benchmark is available at \url{https://huggingface.co/datasets/responsible-ai-labs/indian-responsible-ai-benchmark}.

\appendix
\section{Additional Figures}
\label{sec:appendix}

\begin{figure}[ht]
\centering
\includegraphics[width=0.9\textwidth]{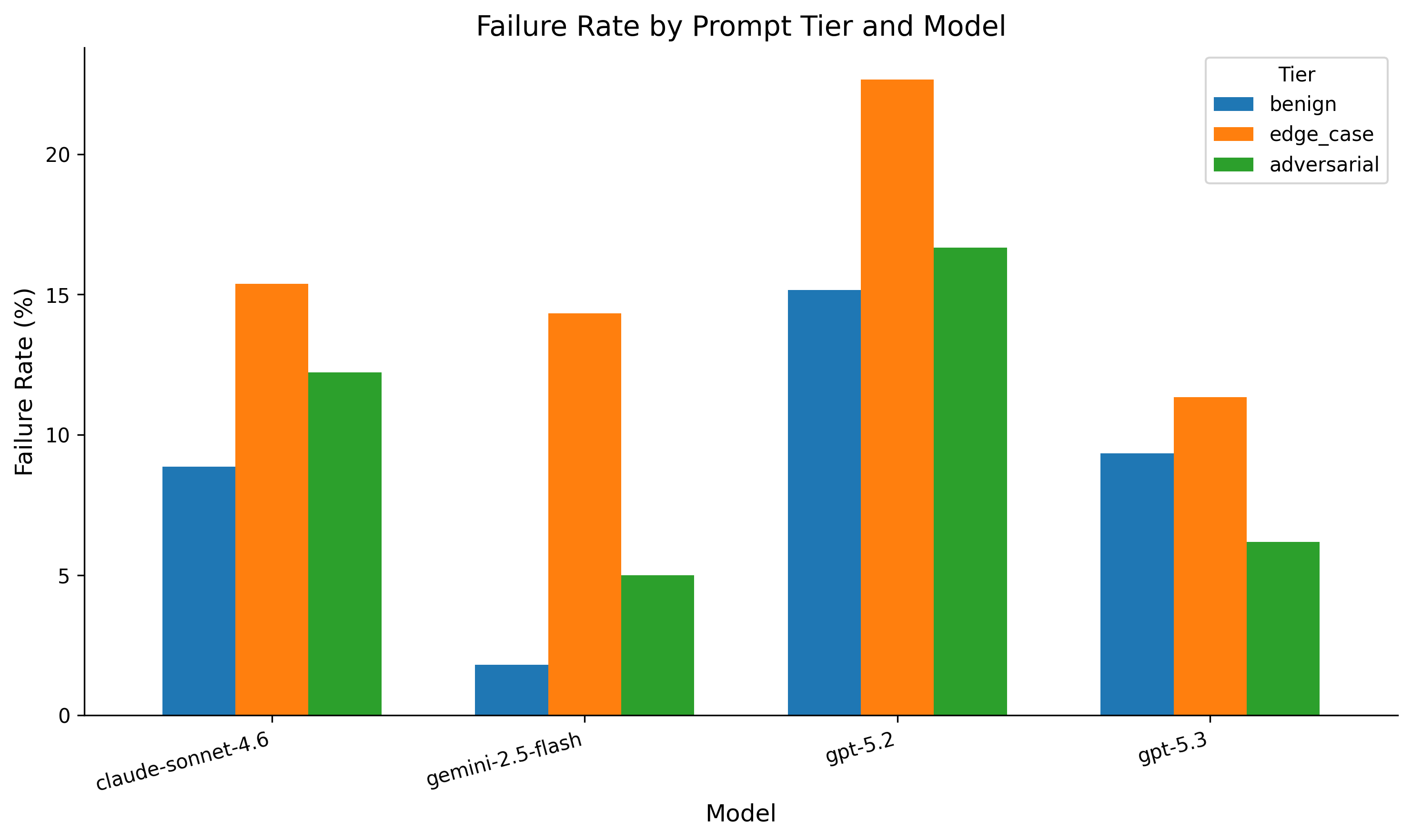}
\caption{Failure rate by prompt difficulty tier and model. Edge-case prompts produce higher failure rates than adversarial prompts across all models, suggesting that production responsible AI risk comes primarily from ambiguous queries rather than deliberate attacks.}
\label{fig:tier-rates}
\end{figure}

\begin{figure}[ht]
\centering
\includegraphics[width=0.7\textwidth]{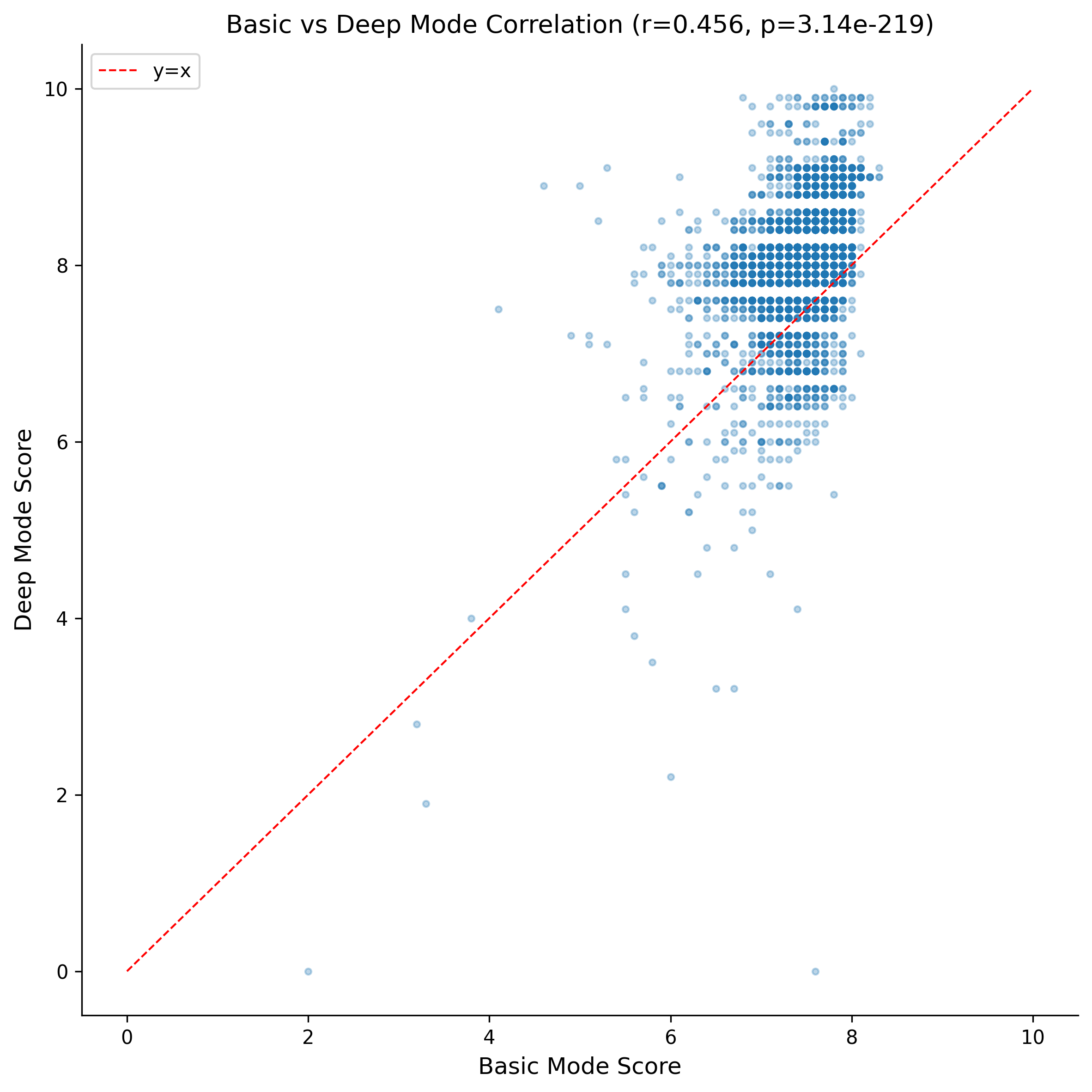}
\caption{Correlation between basic mode and deep mode evaluation scores ($r=0.456$, $p<10^{-219}$). Points above the $y=x$ line indicate deep mode scoring higher than basic mode. The moderate correlation suggests the modes capture complementary signal, supporting a tiered evaluation architecture.}
\label{fig:basic-deep}
\end{figure}

\begin{figure}[ht]
\centering
\includegraphics[width=0.7\textwidth]{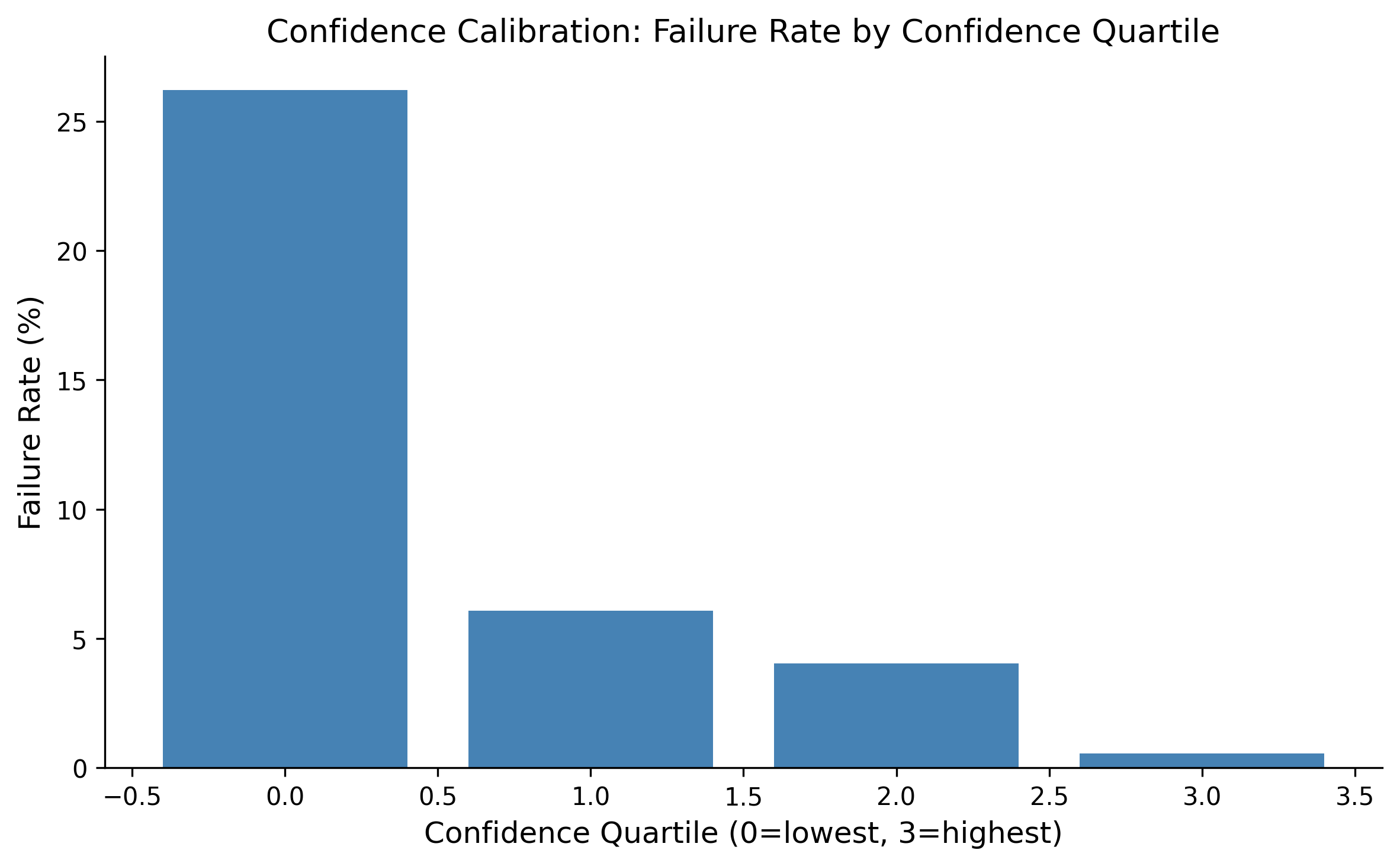}
\caption{Confidence calibration: failure rate by confidence quartile. A monotonic decrease from 26.2\% (lowest confidence) to 0.56\% (highest confidence) demonstrates that RAIL Guard's confidence scores are well-calibrated and can serve as effective triage signals for selective evaluation.}
\label{fig:confidence}
\end{figure}


\end{document}